\documentclass[sigconf]{acmart}
\AtBeginDocument{%
  }

\copyrightyear{2026}
\acmYear{2026}
\setcopyright{cc}
\setcctype{by}
\acmConference[KDD 2026] {Proceedings of the 32nd ACM SIGKDD Conference on Knowledge Discovery and Data Mining V.2}{August 9--13, 2026}{Jeju Island, Republic of Korea.}
\acmBooktitle{Proceedings of the 32nd ACM SIGKDD Conference on Knowledge Discovery and Data Mining V.2 (KDD 2026), August 9--13, 2026, Jeju Island, Republic of Korea}
\acmISBN{979-8-4007-2259-2/2026/08}
\acmDOI{10.1145/3770855.3817790}

\usepackage{hyperref}
\usepackage{url}
\usepackage{tabularray}
\usepackage{color}
\usepackage{multirow}
\usepackage{graphicx}
\usepackage{xcolor}
\usepackage{colortbl}
\usepackage{booktabs}
\usepackage[most,breakable]{tcolorbox}
\usepackage{fancyvrb}
\usepackage{fvextra}
\usepackage{listings}
\usepackage{caption}
\usepackage{fontawesome5}

\begin{document}

% \title{The Name of the Title Is Hope}
% \title{Test-Time Exploration in Unknown Environments}
\title{Test-Time Deep Thinking to Explore Implicit Rules}

\author{Wentong Chen}
\email{cwt\_0139@ruc.edu.cn}
\affiliation{%
  \institution{Renmin University of China}
  \city{Beijing}
  \country{China}
}

\author{Xin Cong}
% \email{}
\affiliation{%
  \institution{Department of Statistics and Data Science, Tsinghua University}
  \city{Beijing}
  \country{China}}

\author{Zhong Zhang}
\affiliation{%
  \institution{School of Computer Science and Engineering, UESTC}
  \city{Chengdu}
  \country{China}
}

\author{Yaxi Lu}
\affiliation{%
 \institution{Department of Computer Science and Technology, Tsinghua University}
 \city{Beijing}
 \country{China}}

\author{Siyuan Zhao}
\affiliation{%
  \institution{School of Mathematical Sciences, Nankai University }
  \city{Tianjin}
  \country{China}}

\author{Yesai Wu}
\affiliation{%
  \institution{Department of Statistics and Data Science, Tsinghua University}
  \city{Beijing}
  \country{China}}

\author{Qinyu Luo}
\affiliation{%
  \institution{Whiting School, Johns Hopkins University}
  \city{Baltimore}
  \state{Maryland}
  \country{USA}}

\author{Haotian Chen}
\authornotemark[1] 
\affiliation{%
  \institution{School of Artificial Intelligence, Shanghai Jiaotong University}
  \city{Shanghai}
  \country{China}}

\author{Yankai Lin}
\authornote{Corresponding author.}
\affiliation{%
  \institution{Renmin University of China}
  \city{Beijing}
  \country{China}}

\author{Zhiyuan Liu}
\affiliation{%
  \institution{Department of Computer Science and Technology, Tsinghua University}
  \city{Beijing}
  \country{China}}

\author{Maosong Sun}
\affiliation{%
  \institution{Department of Computer Science and Technology, Tsinghua University}
  \city{Beijing}
  \country{China}}

\renewcommand{\shortauthors}{Chen et al.}

\begin{abstract}
 With the continuous advancement of Large Language Models (LLMs), intelligent agents are becoming increasingly vital. However, these agents often fail in environments governed by implicit rules—hidden constraints that cannot be observed directly and must be inferred through interaction. 
 This causes agents to fall into repetitive trial-and-error loops, ultimately leading to task failure.
 To address this challenge, we propose \textbf{Test-Time Exploration (TTExplore)},  a framework where a thinker component analyzes interaction history to infer these implicit rules and guide an actor. 
 Effective exploration in this setting critically depends on the reasoning ability of the thinker. However, evaluating deep reasoning trajectories is inherently unstable and difficult, which poses a major obstacle to effective training.
 To overcome this issue, we introduce a novel and stable reinforcement learning pipeline. The core idea is to use accurate task-level scores as indirect rewards to bypass the difficulty of evaluating intermediate reasoning, and to retain only a single thinking node per trajectory to alleviate reward sparsity.
 Using this pipeline, we train a specialized 7B model, \textbf{Exp-Thinker}. 
 Experiments on five text-based embodied tasks show that TTExplore equipped with Exp-Thinker improves baseline agent performance by an average of $14$-$19$ points, demonstrating the effectiveness of explicitly reasoning about implicit rules.
\end{abstract}



  
%     \begin{center}
%         \includegraphics[width=0.85\linewidth]{images/1-new.pdf}
%     \end{center}
%     \caption{Three task examples of the Alfworld environment. We annotate the ``implicit rule'' of each task, which are deduced by humans but the agents can not see. Without understanding these environmental implicit rules, the agent tends to fail.}
%     \label{fig: env_rules}


\maketitle

\section{Introduction}
Large language model (LLM) based agents have demonstrated remarkable capabilities in assisting humans across diverse domains, including deep research~\citep{jin2025search, song2025r1, song2025r1++, chen2026learning}, GUI navigation~\citep{qin2025ui, wu2024atlas, yang2025aria, hong2024cogagent}, and embodied AI~\citep{chang2024agentboard, wang2025ragen, song2024trial}. 
These agents function through iterative interactions with the environment, involving state observation, action generation, and feedback understanding to accomplish tasks. 
As applications expand to open-ended and complex real-world scenarios, enhancing agents' adaptability has become critical. While mainstream research has focused on bolstering agents' inherent general capabilities (e.g., planning and reasoning) via prompt engineering~\citep{yao2023react, shinn2023reflexion} or fine-tuning~\citep{yin2024agent, chen2024agent-flan, xi2024agentgym, hu2025agentgen}, a critical gap remains in how agents effectively adapt to unknown environments.

\begin{figure*}[!t]
\begin{center}
\includegraphics[width=1.0\linewidth]{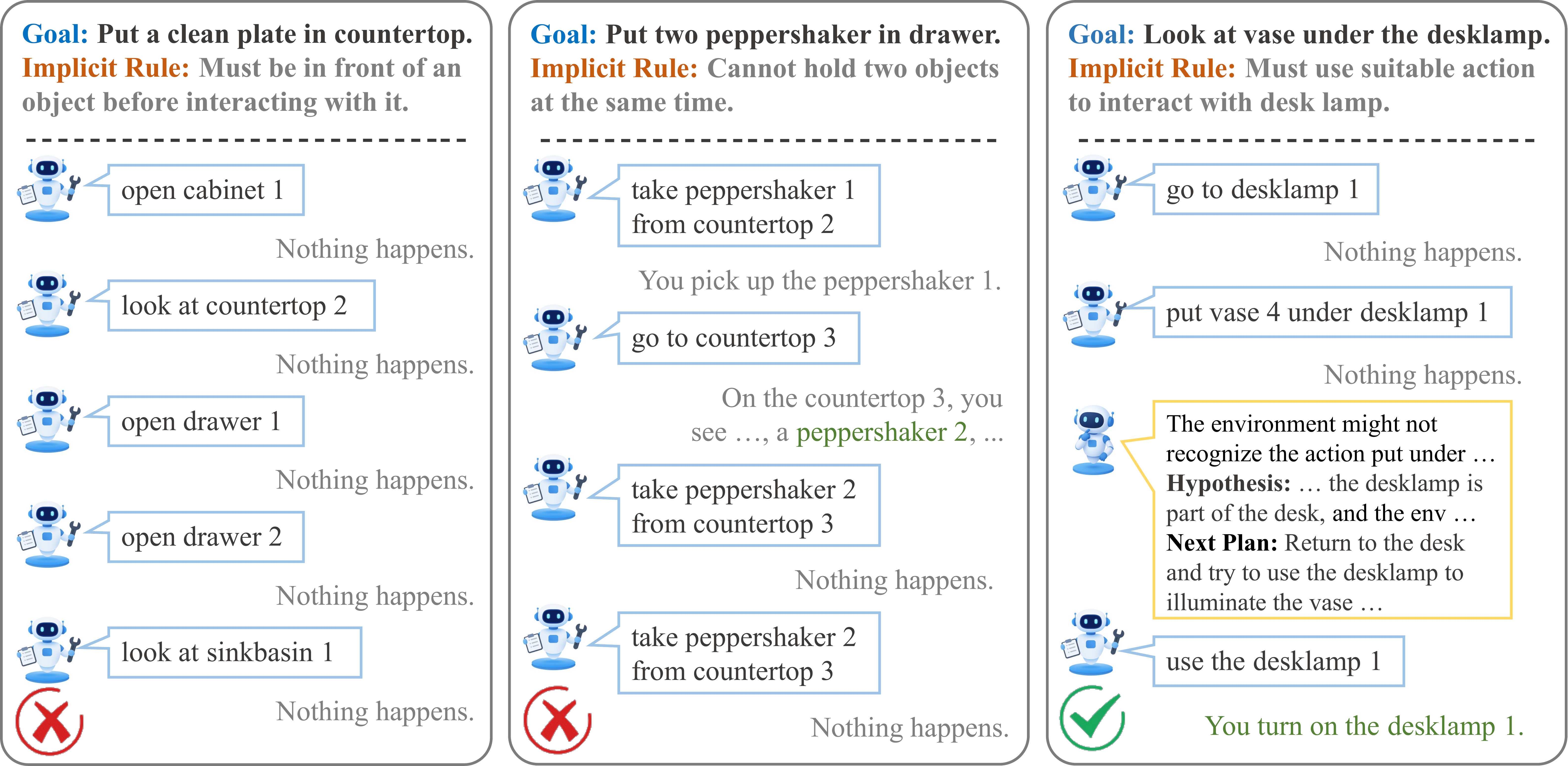}
\end{center}
\caption{Three task examples of the Alfworld environment. We annotate the ``implicit rule'' of each task, which are deduced by humans but the agents can not see. Without understanding these environmental implicit rules, the agent tends to fail.}
\label{fig: env_rules}
\end{figure*}

A key limitation of current methods is their heavy reliance on pre-trained knowledge aligned with the general world, which often fails to capture the latent dynamics of novel environments. 
From a cognitive perspective~\citep{kahneman2011thinking}, explicit information can be handled by System 1–style fast and intuitive reactions, whereas implicit rules correspond to latent environmental dynamics that require System 2–style reasoning and exploration to uncover. We reinterpret these implicit rules from the perspective of \textit{World Model Alignment}: the failure of agents often stems not from a lack of ability, but from acting under a misaligned internal world model.

To illustrate this issue, consider the Alfworld task~\citep{shridhar2020alfworld} shown in Figure~\ref{fig: env_rules}, where agents must navigate household environments governed by implicit constraints, such as “the agent must be in front of an object before interacting with it” or “the agent cannot hold two objects simultaneously.” When unaware of these rules, agents frequently receive uninformative feedback such as “Nothing happened” after attempting invalid actions.

Prior methods such as ReAct~\citep{yao2023react} and Reflexion~\citep{shinn2023reflexion} incorporate planning or reflection mechanisms, yet they still operate within a largely fixed world model and lack a principled way to explore and revise it. As a result, when confronted with feedback that contradicts their internal assumptions—especially in environments with unobservable constraints—agents tend to fall into local exploration loops.
In this view, the core challenge lies in enabling on-the-fly System 2 reasoning for test-time world model updates, allowing agents to progressively align their internal models with the true dynamics of the environment. Such an ability is crucial for effectively solving tasks governed by implicit rules.

To address this challenge, we propose \textbf{Test-Time Exploration (TTExplore)}, a framework designed to enable agents to explore and discover implicit environmental rules during test-time interaction. 
The framework comprises two roles: an actor that generates ReAct-style actions~\citep{yao2023react} and a thinker that performs deep thinking to reason the implicit rules and guide exploration. The thinker monitors task execution by analyzing the actor’s actions and environmental feedback. When recent interactions include failed actions, it summarizes the task trajectory, infers implicit environmental rules, and proposes revised plans, as illustrated in the right panel of Figure~\ref{fig: env_rules}. Throughout task execution, multiple deep thinking steps can be interleaved to regulate the actor’s behavior and facilitate task completion.

The thinker role in our framework is central to exploring environmental knowledge, making its deep thinking ability particularly important. However, improving a model’s thinking ability is highly challenging because we lack direct reward signals to evaluate the quality of its thoughts. A natural idea is to use final task performance (success or failure) as an indirect and sparse reward signal to optimize the thinker role via Reinforcement Learning (RL), such as the GRPO~\citep{shao2024deepseekmath} algorithm. Yet, directly relying on sparse task rewards for RL training is unstable and inefficient. 
To mitigate the credit assignment problem, we design a training strategy that concentrates on key decision points by retaining only a single thinking node per trajectory.
That strategy allows us to more effectively attribute task rewards to key thoughts, enabling us to successfully train a 7B-parameter thinker model, \textbf{Exp-Thinker}.

We evaluate our TTExplore framework with the specially trained thinker model Exp-Thinker on five text-based embodied tasks from Agentboard~\citep{chang2024agentboard}. Results show that our method significantly improves agent performance. Compared with the baseline models \textit{Qwen2.5-7B} and \textit{LLaMA3-8B}, our methods can improve their average performance by approximately $14$ and $19$ points, respectively. 
Furthermore, our method is also effective when combined with the agent training method. Even when the actor model in our framework is well-trained on in-domain tasks, the thinker model can still enhance performance on out-of-domain tasks.

\begin{figure*}[t]
\begin{center}
    \includegraphics[width=1.0\linewidth]{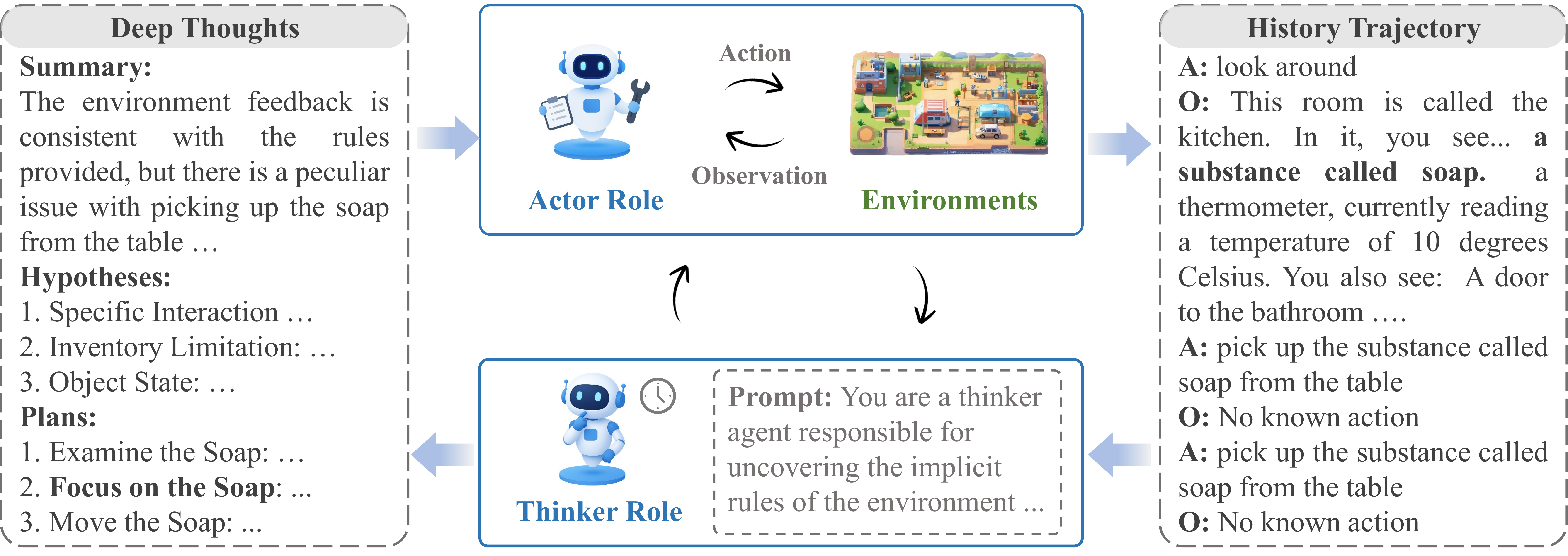}
\end{center}
\caption{The actor role and the thinker role in our TTExplore framework.}
\label{fig: our_framework}
\end{figure*}

\section{Task Formulation} 
Agent tasks that involve multiple rounds of interaction with the environment can be formalized as a Partially Observable Markov Decision Process (POMDP), defined by the tuple $(\mathcal{U}, \mathcal{S}, \mathcal{A}, \mathcal{O}, \mathcal{T})$. Here, $\mathcal{U}$ denotes the instruction space, $\mathcal{S}$ denotes the environment state space, $\mathcal{A}$ denotes the action space, and $\mathcal{O}$ denotes the observation space. The transition function $\mathcal{T}: \mathcal{S}\times\mathcal{A}\to\mathcal{S}$ specifies how the environment state evolves in response to an action. 

For a given task, the agent receives an instruction $u \in \mathcal{U}$, starts from an initial environment state $s_0 \in \mathcal{S}$, and observes an initial partial observation $o_0 \in \mathcal{O}$. At each step $t$, given the previous state $s_{t-1}$ and an action $a_t$, the environment transitions to a new state $s_t$ and produces a partial observation $o_t$. Since the agent $\pi$ cannot access the full environment state $s_t$, it must instead rely on the sequence of historical observations and actions. Typically, an agent $\pi_\theta$ parameterized by $\theta$ predicts the next action $a_{t+1}$ conditioned on the trajectory observed so far, as shown in Eq.~\ref{eq1}:  
\begin{equation}
\label{eq1}
    a_{t+1} \sim \pi_\theta (\cdot \mid u, o_0, a_1, o_1, a_2, o_2, \dots, a_t, o_t).
\end{equation}

This interaction loop continues until the task is completed or a maximum step limit is reached. Finally, a reward $r$ is provided to evaluate task performance over the entire trajectory $traj$. Crucially, in many real-world complex tasks, the observations $\mathcal{O}$ do not explicitly reveal the underlying environment rules. The central challenge for agents is to infer these implicit rules through interaction with the environment, rather than relying solely on surface-level information.

\section{Method}

To enable agents to explore and understand implicit environmental rules during test time, we propose the \textbf{Test-Time Exploration (TTExplore)} framework. At its core is the \emph{thinker} role, a key component that drives effective environment exploration. In this section, we first introduce the overall framework, and then describe our approach to training a specialized thinker model.

\subsection{Test-Time Exploration Framework}

The core principle of our TTExplore framework is to decouple low-level action execution from high-level strategic reasoning. As shown in Figure~\ref{fig: our_framework}, We instantiate this through two distinct roles: an actor $\pi^{actor}$ and a thinker $\pi^{thinker}$.
The actor serves as the primary interaction agent, generating ReAct-style outputs $a_t$ at each step $t$, which consist of a short thought followed by an executable action. In contrast, the thinker functions as a meta-level reasoning agent. It is not invoked at every step; instead, it is triggered when the agent risks getting stuck or requires a deeper understanding of the environment's rules. For simplicity, we activate the thinker at a fixed frequency of $n$ steps in our experiments. Once invoked, the thinker processes the entire trajectory history and produces a deep thought $d_i$, as shown in Eq.~\ref{eq2}.
\begin{equation}
\label{eq2}
    d_i \sim \pi_{\theta}^{thinker} (\cdot|u, o_0, a_1, o_1, a_2, o_2, \dots, a_i, o_i).
\end{equation}

This deep thought typically contains three components: a summary of the current progress, a hypothesis about the latent environmental rules underlying recent failures, and a revised plan for future actions. The output $d_i$ is then prepended to the actor’s subsequent context, as shown in Eq.~\ref{eq3}. In this way, the thinker directly influences the actor’s behavior and helps prevent repeated mistakes.
\begin{equation}
\label{eq3}
    a_{t+1} \sim \pi_\theta^{actor} (\cdot|u, o_0, a_1, o_1, \dots, a_i, o_i, d_i, \dots, a_t, o_t).
\end{equation}

This architecture also enables flexible implementation. The thinker can be the same model as the actor operating under a different prompt, or a larger, more specialized model dedicated to complex reasoning. More details can be seen in Appendix~\ref{append B1}.

\subsection{Reinforcement Learning for Thinker Role}
\label{section-3.2}

\begin{figure*}[t]
\begin{center}
    \includegraphics[width=1.0\linewidth]{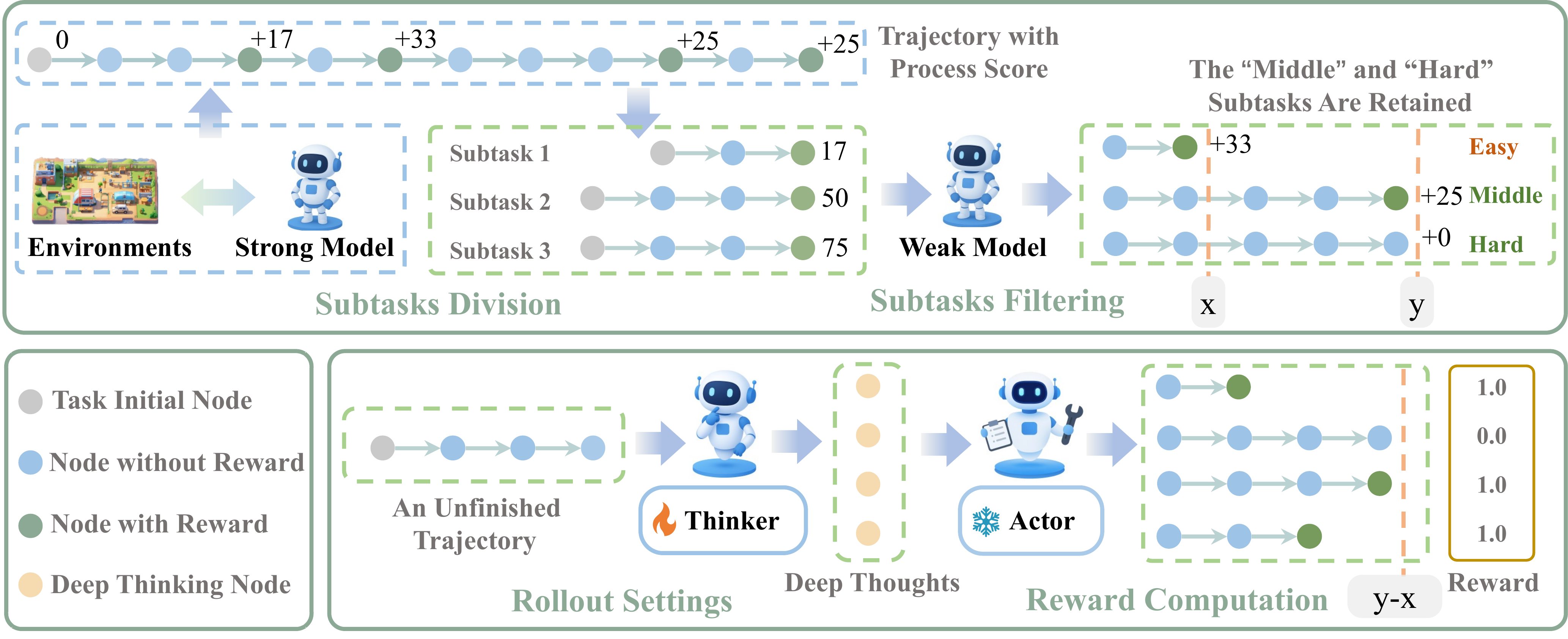}
\end{center}
\caption{The stable training pipeline of our professional thinker model.}
\label{fig: training_pipeline}
\end{figure*}

The main challenge to train the thinker role lies in \textbf{designing a stable and well-defined reward function}. To address this, we adopt an indirect evaluation strategy based on a more stable metric: the task completion performance of the actors. If task performance improves, we consider the thinker’s thoughts beneficial; otherwise, we regard them as detrimental.
Another challenge is how to \textbf{allocate the reward signal to each sampled thought}. In a long trajectory, only one reward is available despite multiple deep-thinking nodes. Our experiments show that sharing the same reward across all nodes makes RL training unstable or prone to collapse, likely due to excessive noise. To mitigate this, we \textbf{retain only a single thinking node per trajectory} during training.

We construct a stable training environment to implement the above designs, as shown in Figure~\ref{fig: training_pipeline}, which consists of four key components: \textbf{\textit{sub-task division, sub-task filtering, rollout settings, and reward computation.}}More details of our training pipeline are shown in Appendix~\ref{append B2}.

\paragraph{Sub-tasks Division} 
To establish sufficiently meaningful and evaluable milestones as the starting points of sub-tasks, we divide the successful trajectories. For each task in our training dataset, we are given a user instruction $u_0$ and an initial environment state $s_0$. Using a strong model $\pi^{strong}$, we generate an execution trajectory $traj^{strong}$ for the task. Based on this trajectory, we partition the task into multiple sub-tasks according to the process score, as defined in Eq.~\ref{eq4}. Each sub-task inherits the initial environment state $s_0$ and the corresponding trajectory $traj_i$ up to the sub-task’s starting point.

In this work, we directly use the provided process scores for sub-task division. For environments with only outcome rewards, alternative strategies—such as manual rules or large model assistance—can be adopted to identify key stages, ensuring the general applicability of our approach.

\begin{equation}
\label{eq4}
    (u_0, s_0) \to traj^{strong} \to [traj_1, traj_2, \dots, traj_n].
\end{equation}

\paragraph{Sub-tasks Filtering} We aim for the thinker model to concentrate on learning solutions for non-trivial and challenging tasks. Therefore, we classify all the sub-tasks into three levels of difficulty (easy, middle, and hard), and exclude those identified as ``easy''. To assess task difficulty, we employ a weak model to generate execution trajectories, which is shown in Eq.~\ref{eq5}. 
First, we set two thresholds, $x$ and $y$, where $0<x<y$. Then, we let the weak model generate $y$-step trajectories from the start state of each sub-task. 
A sub-task is assigned to the easy-level if the weak model completes it within $x$ steps, to the medium-level if completion requires between $x$ and $y$ steps, and to the hard-level otherwise.
\begin{equation}
\label{eq5}
    (u_0, s_0, traj_i) \to traj_i^{weak}.
\end{equation}

\paragraph{Rollout Settings} We reuse the trajectory $traj_i^{weak}$ from the previous step to construct a context that may contain mistaken or unreasonable actions, thus making the intervention of the thinker role necessary.
Before each sampling session, we initialize the environment to the current state of the sub-task by the initial state $s_0$ and two trajectories $traj_i$ and $traj_i^{weak}$. The length of the trajectory $traj_i$ is arbitrary (can be zero), and the length of the trajectory $traj_i^{weak}$ is fixed to $x$. 
Then, the trainable thinker model generates its deep thoughts $m$ times depending on the same historical trajectory, as shown in Eq.~\ref{eq6}. 
\begin{equation}
\label{eq6}
    (u_0, s_0, traj_i, traj_i^{weak}) \to [d_{i,1}, d_{i,2}, ..., d_{i,m}].
\end{equation}

\paragraph{Reward Computation}
Based on the different sampled thoughts, a frozen actor model is invoked to execute a maximum of $(y-x)$ steps, resulting in a total of $m$ trajectories $traj^{actor}$, as shown in Eq.~\ref{eq7}.
\begin{equation}
\label{eq7}
    (u_0, s_0, traj_i, traj_i^{weak}, d_{i,j}) \to traj_{i,j}^{actor} \to r_{i,j}.
\end{equation}

Finally, we will calculate the rewards by the task completion performance of the trajectories: If the process score in a trajectory $traj_{i,j}^{actor}$ improves, the reward $r_{i,j} = 1.0$; otherwise, the reward $r_{i,j} = 0.0$. 
By evaluating a deep thought with one trajectory and giving it a reward $r \in [0, 1]$, we simplify the complex reward assignment problem into a clear, one-to-one causal judgment.

\section{Experiments}

In the experiments section, we first present our experimental setup, including the training and inference settings, datasets, evaluation metrics, and baseline methods. We then analyze the main results, which show that our approaches improve the performance of both base models and well-trained agents. Next, in Sections~\ref{section-4.3}–\ref{section-4.9}, we conduct ablation studies and analytical experiments.

\subsection{Settings}
\label{section-4.1}

% \begin{table}[tb]
% \centering
% \caption{Training hyperparameters for Exp-Thinker.}
% \setlength\tabcolsep{3.5mm}{
% \begin{tabular}{lcc}
% \toprule
% \textbf{Parameter} & \textbf{SFT Stage} & \textbf{RL Stage} \\
% \midrule
% Learning rate & 1e-6 & 1e-6 \\
% Batch size & 32 & 16 \\
% Rollout number & -- & 8 \\
% LR scheduler & cosine & constant \\
% Training epochs & 3 & 3 \\
% BFloat16 & True & True \\
% \bottomrule
% \end{tabular}}
% \label{tab: training params}
% \end{table}

\paragraph{Training Settings} 
Our main work is to train a professional model \textbf{Exp-Thinker} as the thinker role in our \textbf{TTExplore} framework. We choose the \textit{Qwen2.5-7B-Instruct}~\citep{hui2024qwen2} as the backbone, and use Supervised Fine-Tuning (SFT) and Reinforcement Learning (RL) to train it. 
For the SFT stage, we employ the TRL training framework \citep{vonwerra2022trl} and train the model for 3 epochs on $10,797$ training samples.  
For the RL stage, we adopt the Verl framework \citep{sheng2024verl}, utilize $6,232$ sub-tasks, and train for multiple epochs until model convergence.  
% Detailed hyperparameters for both stages are provided in Table~\ref{tab: training params}.

% We also train actor models for further experiments, which are based on \textit{Qwen2.5-7B-Instruct} and \textit{LLaMA3-8B-Instruct}~\citep{dubey2024llama}, respectively. 
% More details can be seen in Appendix~\ref{append C1}.

We further fine-tune the \textit{LLaMA3-8B-Instruct}~\citep{dubey2024llama} and \textit{Qwen2.5-7B-Instruct} models to obtain improved actor agents, denoted as \textbf{LLaMA3-Actor} and \textbf{Qwen2.5-Actor}, respectively. Both models are trained using reinforcement learning on $2,509$ tasks from the Alfworld and Sciworld environments. For LLaMA3-Actor, we incorporate additional SFT samples from AgentGym~\citep{xi2024agentgym} to facilitate cold-start initialization. Note that we exclude the BabyAI task samples from the AgentGym dataset to ensure that BabyAI remains an out-of-domain evaluation environment.

\paragraph{Inference Details}  
All experiments are conducted using vLLM~\citep{kwon2023vllm} as the inference engine. The default temperature is fixed at $0.0$, corresponding to greedy decoding. For all test tasks, we limit the maximum number of interaction steps to $50$. We trigger the thinker at a fixed frequency, which means after every $n$ steps of interaction, the thinker is activated. In our experiments, we set $n = 6$.  

\paragraph{Datasets and Metrics}
We evaluate our methods on five text-based embodied tasks, including ALFworld~\citep{shridhar2020alfworld}, Sciworld~\citep{wang2022scienceworld}, BabyAI~\citep{chevalier2018babyai}, PDDL~\citep{vallati2015pddl}, and Jericho~\citep{hausknecht2020jericho}. We follow the metrics, few-shot examples, and action space descriptions in Agentboard~\citep{chang2024agentboard}. We apply \textbf{Process Score} as the metric, which is in the range $0.0$ to $100.0$. Specifically, we classified the Alfworld and Sciworld as in-domain tasks, and the other three as out-of-domain tasks. 
We show the number of tasks in different datasets in Table~\ref{tab: task number}. The training tasks are used for training LLaMA3-Actor and Qwen2.5-Actor, and the sub-tasks are used for training the Exp-Thinker.

\begin{table}[tb]
  % \scriptsize
  % \small
  \centering
  \caption{The number of tasks in different datasets.}
  \setlength\tabcolsep{0.6mm}{
  \begin{tabular}{lccccc}
    \toprule
    & \textbf{Alfworld} & \textbf{Sciworld} & \textbf{BabyAI} & \textbf{Jericho} & \textbf{PDDL} \\
    \midrule
    Training tasks & 800 & 1,709 & -- & -- & -- \\
    Training subtasks & 1,093 & 5,139 & -- & -- & -- \\ 
    Evaluation tasks & 134 & 90 & 112 & 20 & 60 \\
    \bottomrule
  \end{tabular}}
  \label{tab: task number}
\end{table}

\paragraph{Baselines}
We compared our method with large models and fine-tuned agents: (1) \textbf{Large Models}, including GPT-4o~\citep{hurst2024gpt} and Qwen2.5-72B-Instruct~\citep{hui2024qwen2}; (2) \textbf{Fine-Tuned Agents}, including Agent-FLAN~\citep{chen2024agent-flan}, AgentGym~\citep{xi2024agentgym}, and AgentRefine~\citep{fu2025agentrefine}.
% In our experiments, we use \textit{LLaMA3-8B-Instruct}, \textit{Qwen2.5-7B-Instruct}, and their continue fine-tuned agents (LLaMA3-Actor and Qwen2.5-Actor) as the \textit{actor} role. Then, we use our specially trained model \textbf{Exp-Thinker} as the \textit{thinker} role.

\begin{table*}[tb]
  \centering
  \caption{\textbf{Main Results of Our Methods.} The first column and second column list the \textit{Actor} role and the \textit{Thinker} role in our proposed \textbf{TTExplore} framework, respectively. If the \textit{Thinker} role is \textit{No}, the framework back to the \textbf{ReAct} method, where the actor produces short thoughts. \textbf{Exp-Thinker} is trained specifically for the thinker. The terms "Succ." and "Proc." abbreviate "success rate" and "process score", respectively. The last column reports the overall average process score across tasks. In-domain tasks are visually distinguished with a \colorbox{green!10}{green} background for each agent system.}
  \setlength\tabcolsep{2.35mm}{
  \begin{tabular}{llccccccccccc}
    \toprule
    \multicolumn{2}{c}{\textbf{Agent Systems}} & \multicolumn{2}{c}{Alfworld} & \multicolumn{2}{c}{Sciworld} & \multicolumn{2}{c}{BabyAI} & \multicolumn{2}{c}{Jericho} &  \multicolumn{2}{c}{PDDL} & \textbf{Mean} \\
    \cmidrule(lr){1-2}
    \cmidrule(lr){3-4}
    \cmidrule(lr){5-6} 
    \cmidrule(lr){7-8}
    \cmidrule(lr){9-10}
    \cmidrule(lr){11-12}  
     \textbf{Actor} & \textbf{Thinker}  & Succ. & Proc. & Succ. & Proc.  & Succ. & Proc. & Succ. & Proc. & Succ. & Proc. & \textbf{Proc.} \\
    \midrule
    \multicolumn{13}{c}{\cellcolor{gray!10}\textit{Large Models \& Fine-Tuned Agents}} \\
    \midrule
    GPT-4o~\citep{hurst2024gpt} & No & 66.40 & 79.90 & 40.00 & 76.90 & 48.20 & 64.10 & 10.00 & 34.00 & 61.70 & 69.80 & 	64.94 \\
    Qwen2.5-72B~\citep{hui2024qwen2} & No & 91.79 & 93.84 & 68.88 & 76.74 & 55.35 & 70.07 & 15.00 & 30.87 & 56.66 & 70.50 & \textbf{68.40} \\ 
    
    \midrule
    Agent-FLAN~\citep{chen2024agent-flan} & No & \cellcolor{green!10}67.20 & \cellcolor{green!10}79.70 & \ \ 1.10 & 10.90 & 25.00 & 35.30 & \ \ 0.00 & 10.10 &\ \ 8.30 & 25.50 & 32.30 \\ 
    AgentGym~\citep{xi2024agentgym} & No & \cellcolor{green!10}61.90 & \cellcolor{green!10}76.90 & \cellcolor{green!10}18.90 & \cellcolor{green!10}47.50 & \cellcolor{green!10}47.30 & \cellcolor{green!10}61.40 & \ \ 0.00 & 12.90 & \ \ 1.70 & 16.60 & 43.06 \\
    AgentRefine~\citep{fu2025agentrefine} & No & 44.80 & 63.80 & 14.40 & 42.60 & 37.50 & 50.40 & 10.00 & 32.30 & 16.60 & 37.80 & \textbf{45.38} \\ 

    \midrule
    \multicolumn{13}{c}{\cellcolor{blue!7}\textit{Our Methods}} \\
    \midrule
    
    \multirow{2}{*}{LLaMA3-8B~\citep{dubey2024llama}} 
    & No  & \ \ 7.46 & 32.15 & 13.33 & 20.31 & 26.78 & 42.44 & 0.00 & 14.23 & 10.00 & 29.94 & 27.81 \\
    & Exp-Thinker & \cellcolor{green!10}32.08 & \cellcolor{green!10}63.49 & \cellcolor{green!10}44.44 & \cellcolor{green!10}55.43 & 41.96 & 58.15 & 5.00 & 26.31 & 16.66 & 30.06 & \textbf{46.69} \\
    \midrule

    \multirow{2}{*}{Qwen2.5-7B~\citep{hui2024qwen2}} 
    & No & 68.65 & 80.03 & 33.33 & 39.82 & 37.50 & 50.16 & 0.00 & 8.95  & 18.33 & 25.38 & 40.87 \\
    & Exp-Thinker & \cellcolor{green!10}90.29 & \cellcolor{green!10}94.96 & \cellcolor{green!10}65.55 & \cellcolor{green!10}72.69 & 50.00 & 63.12 & 5.00 & 19.94 & 8.33  & 20.87 & \textbf{54.32} \\
    \midrule
    
    \multirow{2}{*}{LLaMA3-Actor}
    & No  & \cellcolor{green!10}68.66 & \cellcolor{green!10}78.92 & \cellcolor{green!10}45.56 & \cellcolor{green!10}54.87 & 23.21 & 35.55 & 0.00 & 15.98 & 5.00  & 20.19 & 41.10 \\
    & Exp-Thinker & \cellcolor{green!10}77.61 & \cellcolor{green!10}87.75 & \cellcolor{green!10}51.11 & \cellcolor{green!10}63.14 & 35.71 & 46.41 & 0.00 & 17.11 & 10.00 & 32.11 & \textbf{49.30} \\
    \midrule
    
    \multirow{2}{*}{Qwen2.5-Actor}
    & No & \cellcolor{green!10}94.77 & \cellcolor{green!10}97.76 & \cellcolor{green!10}77.77 & \cellcolor{green!10}83.07 & 40.17 & 50.62 & 0.00 & 17.44 & 15.00 & 29.76 & 55.73 \\
    & Exp-Thinker & \cellcolor{green!10}97.76 & \cellcolor{green!10}98.50 & \cellcolor{green!10}74.44 & \cellcolor{green!10}82.27 & 48.21 & 60.25 & 0.00 & 17.64 & 11.66 & 31.43 & \textbf{58.02} \\
    \bottomrule
  \end{tabular}}
  \label{tab: main-results}
\end{table*}

\subsection{Main Results}
\label{section-4.2}

\paragraph{(1) The TTExplore framework with Exp-thinker can enhance the performance of baseline models.}
Our method yields substantial improvements for base models across both in-domain and out-of-domain tasks. When employing LLaMA3-8B and Qwen2.5-7B as the actor, our approach consistently outperforms the baseline setting (without a thinker) on nearly all tasks, achieving average performance gains of $14$-$19$ points. For instance, with LLaMA3-8B on in-domain tasks, the thinker model Exp-Thinker boosts performance from $32.15$ and $20.31$ to $63.49$ and $55.43$, which is an increase of more than $30$ points. On out-of-domain tasks, the improvements remain notable, averaging nearly $10$ points. These gains elevate the performance of the base model LLaMA3-8B to a level comparable with fine-tuned agents such as AgentRefine. Overall, these results highlight both the effectiveness and the strong generalization capabilities of our framework and the Exp-Thinker.

\begin{figure*}[t]
\begin{center}
    \includegraphics[width=\linewidth]{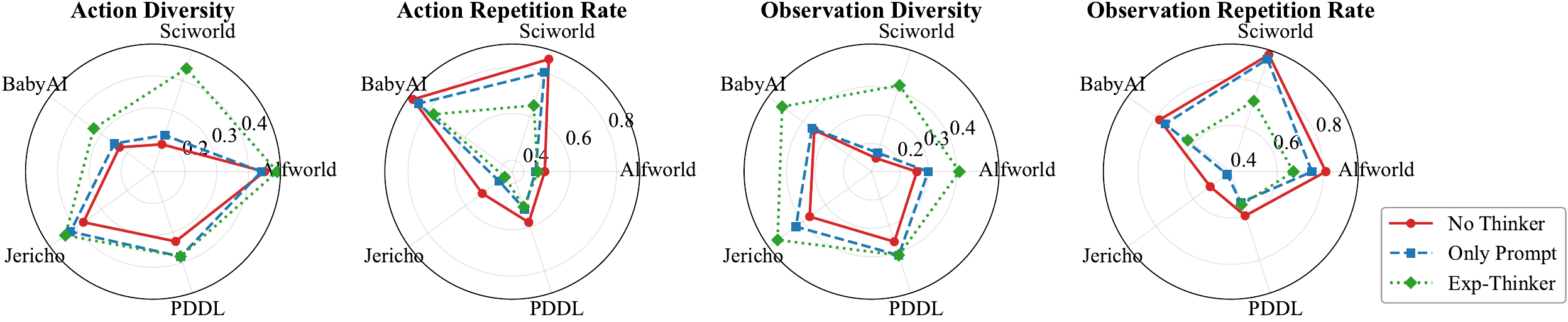}
\end{center}
\caption{The quantitative analysis of exploration behavior for different methods.}
\label{fig: llama_radar}
\end{figure*}

% \noindent\textbf{(2) The Exp-thinker can enhance the performance of well-trained agents on out-of-domain tasks.}
\paragraph{(2) The Exp-thinker can enhance the performance of well-trained agents on out-of-domain tasks.}
Prior studies have improved agent performance primarily through data construction for SFT or RL. Building on this, we further train LLaMA3-8B and Qwen2.5-7B on in-domain tasks, yielding stronger actor variants, LLaMA3-Actor and Qwen2.5-Actor. For LLaMA3-Actor, incorporating the thinker model improves performance across all tasks, with an average gain of $8$ points. Qwen2.5-Actor, a well-trained agent on in-domain tasks, achieves the score of $97.76$ and $83.07$ on Alfworld and Sciworld, respectively. Even for such a strong agent, our method still enhances its out-of-domain performance, most notably on BabyAI, where accuracy increases from $50.62$ to $60.25$. 
This demonstrates that the traditional agent abilities improved by SFT and RL are complementary to the ``test-time exploration'' capability provided by our thinker. Our approach is not a replacement for existing training methods, but rather a powerful complement.

\subsection{Does the Thinker Role Improve the Actor's Exploration Behavior?}
\label{section-4.3}
% 回到论文主要目标
This section provides a quantitative analysis of how our framework enhances the model’s exploratory behavior.
% 提出评价指标
In complex environmental interaction tasks, agents often fail due to repetitive behaviors or overly localized exploration. To evaluate exploration more systematically, we introduce four metrics: \textbf{\textit{action diversity, action repetition rate, observation diversity, and observation repetition rate}}.
Given an execution trajectory $T(a_1, o_1, a_2, o_2, \dots, a_n, o_n)$, \textbf{action diversity} is defined as the number of distinct actions divided by the trajectory length, while \textbf{action repetition rate} is defined as the proportion of top-$k$ most frequent actions within the trajectory. Similarly, \textbf{observation diversity} measures the ratio of distinct observations visited to the trajectory length, and \textbf{observation repetition rate} measures the proportion of the top-$k$ most frequently visited observations. In our experiments, we set $k=3$. These four metrics allow us to quantify the exploratory behavior of a single trajectory. The higher diversity and lower repetition rate mean better exploratory behavior.

\paragraph{Results.}
We compare the exploratory behaviors of \textit{LLaMA-8B-Instruct} under different methods, as illustrated in Figure~\ref{fig: llama_radar}. 
The results indicate that incorporating the Exp-Thinker consistently increases both action and observation diversity while reducing repetition rates, demonstrating its effectiveness in enhancing exploration. These findings provide direct evidence for our central hypothesis: the thinker mitigates the ``local exploration loops'' and ``repetitive trial-and-error'' behaviors identified in the introduction. By encouraging more diverse actions and observations, the agent is steered away from unproductive cycles, leading to the performance gains reported in Table~\ref{tab: main-results}.

\subsection{Comparison of Different Thinker Roles}
\label{section-4.4}

We evaluate our \textbf{TTExplore} framework by incorporating different thinker roles. In our experiments, we adopt \textit{LLaMA3-8B-Instruct} and \textit{Qwen2.5-7B-Instruct} as the actor roles. For the thinker roles, we consider two categories:
(1) \textbf{Base models}, including \textit{Qwen2.5-7B-Instruct}, \textit{Qwen2.5-72B-Instruct}, and \textit{Qwen3-8B} (thinking mode).
(2) \textbf{Trained thinker models}, obtained through different training strategies: \textit{Only SFT}, \textit{Only RL}, and \textit{SFT followed by RL}. 

\begin{table*}[tb]
  % \scriptsize
  % \small
  \centering
  \caption{The performance of different base models and specifically trained thinkers as the thinker role. In-domain tasks are visually distinguished with a \colorbox{green!10}{green} background for each agent system.}
  \setlength\tabcolsep{2.0mm}{
  \begin{tabular}{llccccccccccc}
    \toprule
    \multicolumn{2}{c}{\textbf{Agent Systems}} & \multicolumn{2}{c}{Alfworld} & \multicolumn{2}{c}{Sciworld} & \multicolumn{2}{c}{BabyAI} & \multicolumn{2}{c}{Jericho} &  \multicolumn{2}{c}{PDDL} & \textbf{Mean} \\
    \cmidrule(lr){1-2}
    \cmidrule(lr){3-4}
    \cmidrule(lr){5-6} 
    \cmidrule(lr){7-8}
    \cmidrule(lr){9-10}
    \cmidrule(lr){11-12}  
     \textbf{Actor} & \textbf{Thinker}  & Succ. & Proc. & Succ. & Proc.  & Succ. & Proc. & Succ. & Proc. & Succ. & Proc. & \textbf{Proc.} \\
    \midrule
    \multirow{7}{*}{LLaMA3-8B}
    & No & 7.46 & 32.15 & 13.33 & 20.31 & 26.78 & 42.44 & 0.00 & 14.23 & 10.00 & 29.94 & 27.81 \\
    & Qwen2.5-7B & 9.70 & 39.17 & 10.00 & 17.69 & 27.67 & 42.02 & 5.00 & 26.70 & 5.00 & 26.01 & 30.32  \\
    & Qwen3-8B & 20.15 & 53.42 & 25.56 & 36.47 & 26.79 & 40.01 & 10.00 & 35.42 & 15.00 & 35.76 & \textbf{40.22}  \\
    & Qwen2.5-72B & 13.43 & 48.13 & 11.11 & 21.32 & 27.67 & 41.72 & 0.00 & 25.97 & 16.66 & 36.05 & 34.64 \\
    
    \cmidrule(lr){2-13}
    & Exp-Thinker (Only SFT) & \cellcolor{green!10}11.94 & \cellcolor{green!10}47.13 & \cellcolor{green!10}24.44 & \cellcolor{green!10}33.15 & 25.00 & 38.88 & 0.00 & 23.19 & 6.66 & 28.58 & 34.19 \\
    & Exp-Thinker (Only RL) & \cellcolor{green!10}14.92 & \cellcolor{green!10}48.00 & \cellcolor{green!10}23.33 & \cellcolor{green!10}35.59 & 27.67 & 44.33 & 0.00 & 21.43 & 5.00 & 25.54 & 34.98  \\
    & Exp-Thinker (SFT + RL) & \cellcolor{green!10}32.08 & \cellcolor{green!10}63.49 & \cellcolor{green!10}44.44 & \cellcolor{green!10}55.43 & 41.96 & 58.15 & 5.00 & 26.31 & 16.66 & 30.06 & \textbf{46.69} \\
    
    \midrule
    \multirow{7}{*}{Qwen2.5-7B} 
    & No & 68.65 & 80.03 & 33.33 & 39.82 & 37.50 & 50.16 & 0.00 & 8.95 & 18.33 & 25.38 & 40.87 \\
    & Qwen2.5-7B & 57.46 & 76.18 & 48.88 & 54.61 & 49.10 & 63.27 & 0.00 & 11.93 & 6.66 & 21.08 & 45.41 \\
    & Qwen3-8B & 77.61 & 89.18 & 48.89 & 55.76 & 50.00 & 64.32 & 0.00 & 19.43 & 8.33 & 30.71 & 51.88 \\
    & Qwen2.5-72B & 83.58 & 92.91 & 60.00 & 69.20 & 58.03 & 72.08 & 5.00 & 20.57 & 10.00 & 28.76 & \textbf{56.70} \\

    \cmidrule(lr){2-13}
    & Exp-Thinker (Only SFT) & \cellcolor{green!10}80.59 & \cellcolor{green!10}90.79 & \cellcolor{green!10}62.22 & \cellcolor{green!10}70.82 & 50.00 & 64.41 & 0.00 & 18.18 & 5.00 & 21.33 & 53.11 \\
    & Exp-Thinker (Only RL) & \cellcolor{green!10}69.40 & \cellcolor{green!10}83.27 & \cellcolor{green!10}56.66 & \cellcolor{green!10}66.70 & 55.35 & 69.55 & 0.00 & 10.59 & 8.33 & 21.12 & 50.25 \\
    & Exp-Thinker (SFT + RL) & \cellcolor{green!10}90.29 & \cellcolor{green!10}94.96 & \cellcolor{green!10}65.55 & \cellcolor{green!10}72.69 & 50.00 & 63.12 & 5.00 & 19.94 & 8.33 & 20.87 & \textbf{54.32} \\
    \bottomrule
  \end{tabular}}
  \label{tab: different-train}
\end{table*}

\paragraph{Q1: Can we just use a large off-the-shelf model for the thinker role?}
Results in Table~\ref{tab: different-train} reveal a crucial insight: While the larger model Qwen2.5-72B generally serves as a better thinker than its smaller counterparts, our specialized thinker Exp-Thinker (SFT + RL) consistently matches or even surpasses the 72B-parameter model in most tasks. This demonstrates that targeted training on reasoning about environmental interactions is more effective than simply relying on the scale of a general-purpose model. It validates the necessity of our proposed training pipeline.

\paragraph{Q2: Why is the ``SFT + RL'' training pipeline the best approach?}
Our ablation confirms the complementary roles of supervised fine-tuning (SFT) and reinforcement learning (RL). ``Only SFT'' offers a strong initialization but is limited by the static dataset, while ``Only RL'' struggles with a weak starting state. The combined ``SFT + RL'' approach proves most effective, using SFT to overcome cold-start challenges and reinforcement learning to further enhance the thinker's reasoning abilities in an interactive environment, allowing it to discover strategies beyond the initial supervision.

\subsection{How to Choose an Optimal Thinking Frequency During Inference?}
\label{section-4.5}

\begin{figure}[t]
\begin{center}
    \includegraphics[width=\linewidth]{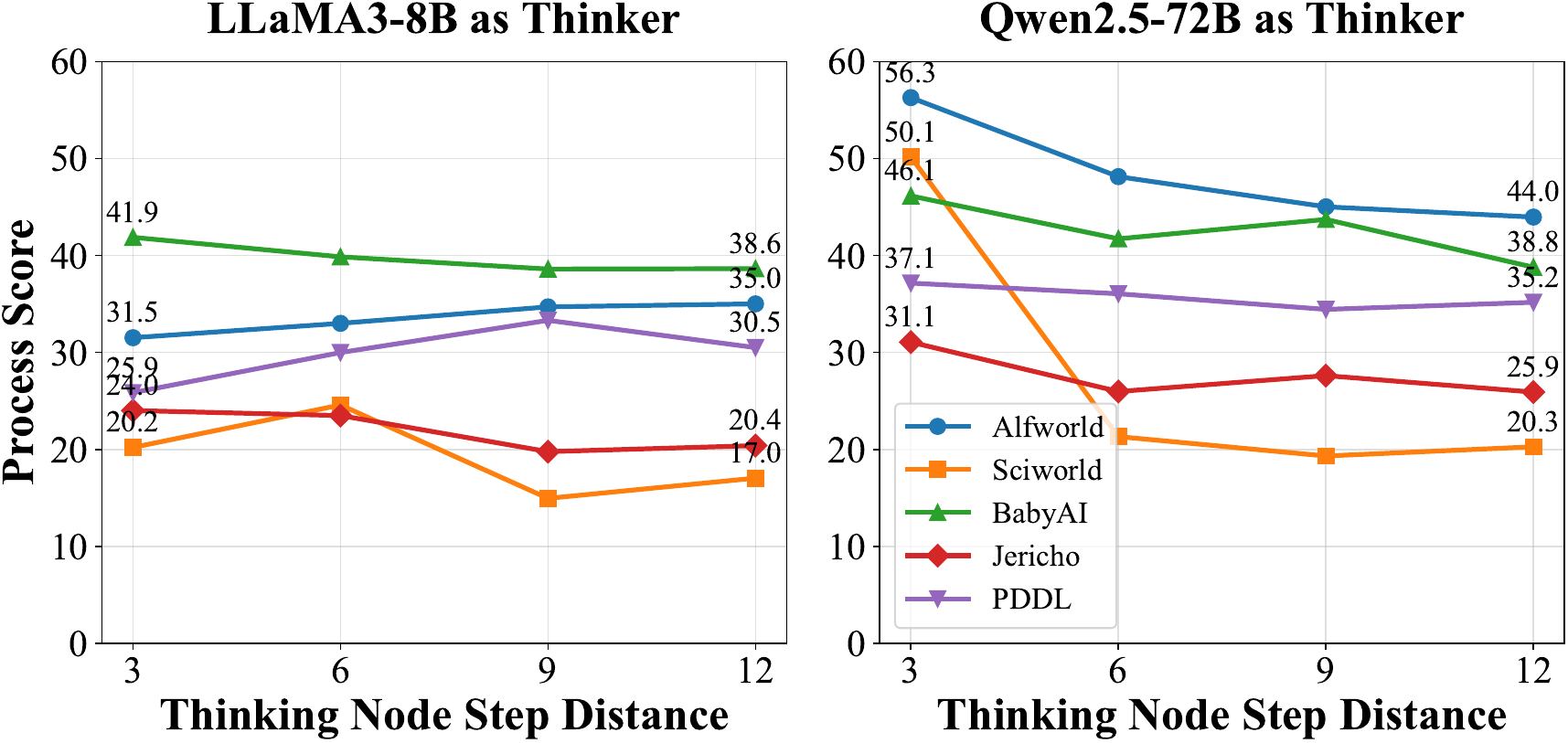}
\end{center}
\caption{Performance at different deep thinking frequencies. A larger step distance indicates a lower frequency.}
\label{fig: think_freq}
\end{figure}

We investigate how different deep thinking frequencies affect task performance during inference. In this experiment, we adopt \textit{LLaMA3-8B-Instruct} as the actor, paired either with itself or with a larger model, \textit{Qwen2.5-72B-Instruct}, as the thinker. The thinker is invoked at fixed intervals of $3$, $6$, $9$, and $12$ steps.

\paragraph{Results.}
As shown in Figure~\ref{fig: think_freq}, the optimal deep thinking interval varies across tasks. On average, when using a smaller thinker model, invoking thinking every $6$ steps yields better performance, whereas with a larger thinker, a higher thinking frequency—specifically every $3$ steps—proves most effective.
We attribute this discrepancy primarily to differences in thinking quality. High-quality thinking can provide more reliable guidance for problem-solving and exploration, while lower-quality thinking may disrupt the actor’s execution flow. Moreover, increasing the thinking frequency inevitably incurs higher inference-time costs. Considering the trade-off between performance and efficiency, we adopt a thinking interval of $6$ steps in our experimental setup.
Nevertheless, this fixed choice may not be optimal for all tasks or model configurations. While this work focuses on increasing the quality of thinking nodes, an important direction for future research is to develop mechanisms for dynamically adjusting thinking frequency during inference.

% \subsection{Time Cost of Different Methods}
\subsection{Comparison of Test-Time Scaling Methods}
\label{section-4.6}

\begin{table*}[t]
\centering
    \caption{Comparison of time cost and performance across methods. The notation "Reflexion ($\times$ N)" denotes a maximum of $N$ reflections and retry iterations. The score refers to the process score of the task, and the unit of time is seconds.}
  \setlength\tabcolsep{2.0mm}{
    \begin{tabular}{lrrrrrrrrrrrrr}
    \toprule
    \multirow{2}{*}{\textbf{Methods}} & \multicolumn{2}{c}{Alfworld} & \multicolumn{2}{c}{Sciworld} & \multicolumn{2}{c}{BabyAI} & \multicolumn{2}{c}{Jericho} & \multicolumn{2}{c}{PDDL} &
    \textbf{Mean} & \multicolumn{2}{c}{\textbf{Total Time}} \\
    \cmidrule(lr){2-3}
    \cmidrule(lr){4-5}
    \cmidrule(lr){6-7}
    \cmidrule(lr){8-9}
    \cmidrule(lr){10-11}
    \cmidrule(lr){13-14}
    & Score & Time & Score & Time & Score & Time & Score & Time & Score & Time & \textbf{Score} & \textbf{Cost} & \textbf{Ratio} \\
    \midrule
    ReAct      & 78.9 & 71  & 33.4 & 95  & 49.3 & 91  & 7.3  & 53  & 23.1 & 79  & 38.4 & \textbf{389}  & 1.0 \\
    TTExplore       & 91.9 & 111 & 75.9 & 116 & 69.4 & 130 & 19.2 & 99  & 21.4 & 104 & \cellcolor{orange!10}\textbf{55.5} & \cellcolor{orange!10} 559 & \cellcolor{orange!10} 1.4 \\

    \midrule
    Reflexion ($\times$ N)  & 85.6 & 341 & 56.9 & 436 & 75.6 & 469 & 18.0 & 291 & 38.3 & 301 & 54.9 & 1838 & 4.7 \\
    Best-of-N   & 96.3 & 357 & 70.5 & 443 & 84.2 & 467 & 26.7 & 295 & 54.0 & 362 & 66.3 & 1925 & 5.0 \\
    TTExplore + Best-of-N  & 99.4 & 589 & 91.1 & 606 & 85.3 & 687 & 31.5 & 507 & 43.8 & 519 & \cellcolor{orange!10}\textbf{70.2} & \cellcolor{orange!10} 2907 & \cellcolor{orange!10} 7.5 \\
    \bottomrule
    \end{tabular}}
  \label{tab: time-cost}
\end{table*}

In this section, we compare our TTExplore framework with standard \textit{ReAct} and two representative test-time scaling methods: \textit{Reflexion} and \textit{Best-of-N}. Our method is orthogonal to Best-of-N, we additionally include a combined variant that applies Best-of-N sampling on top of TTExplore to illustrate their complementarity.
In our experiments, we use \textit{Qwen2.5-7B-Instruct} as the actor role and the trained Exp-Thinker as the thinker role. For Reflexion and Best-of-N, we fix the number of samples to $5$ to maintain a relatively fair computational comparison across all methods.

\paragraph{Results.}
As shown in Table~\ref{tab: time-cost}, our TTExplore framework incurs a moderate computational overhead, running approximately $1.4\times$ slower than standard ReAct. However, this overhead is substantially lower than that of Reflexion ($4.7\times$) and Best-of-N sampling ($5.0\times$). Importantly, TTExplore remains fully compatible with Best-of-N sampling, enabling users to trade additional computational budget for further performance gains. When Best-of-N sampling is applied on top of TTExplore (TTExplore + Best-of-N), the combined approach achieves the strongest overall performance.
% more analysis
Understanding the computational overhead introduced by our TTExplore method is essential for assessing its practical utility. Although our training process involves multiple sampling-and-evaluation iterations, our inference-time procedure uses only a single sample. Unlike traditional test-time scaling methods that rely heavily on multiple samples (e.g., Best-of-N or Reflexion), our method integrates deep thinking results directly into a single rollout. While the \textbf{thinker role} may consider multiple possible solution directions during deep thinking, the \textbf{actor role} performs actions sequentially, \textbf{without branching or backtracking}. This design is consistent with how humans explore real environments, where tasks are irreversible, and exploration must occur within a single action budget.

\begin{figure}[t]
\begin{center}
    \includegraphics[width=0.7\linewidth]{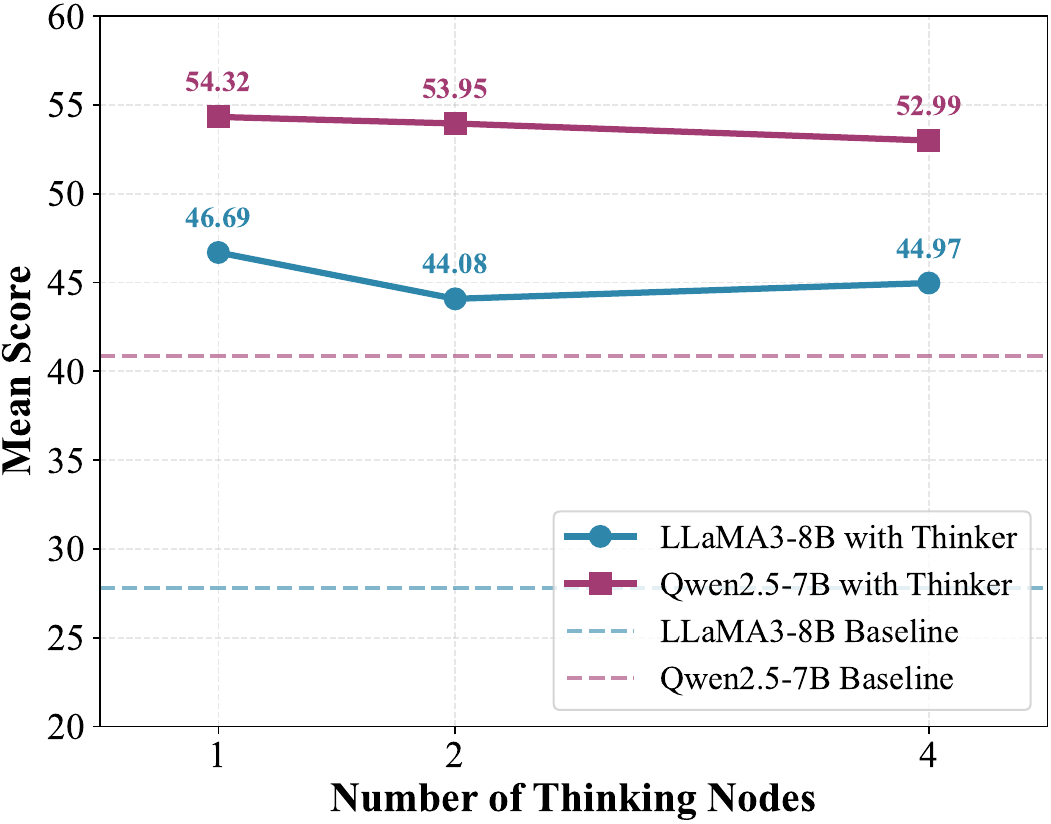}
\end{center}
\caption{Performance at thinker models that have different thinking nodes per trajectory during training.}
\label{fig: think-node-per-traj}
\end{figure}

\subsection{Is the Single Thinking Node per Trajectory Necessary for Training?}
\label{section-4.7}

In this section, we examine how the number of deep thinking nodes within each trajectory influences the stability of RL training. The trajectory-level reward, computed using the GRPO~\citep{shao2024deepseekmath} algorithm, is uniformly assigned to all thinking nodes within the same trajectory. Under our initial configuration, trajectories containing more than six thinking nodes consistently lead to training collapse after several dozen to approximately one hundred training iterations.
Based on this observation, we restrict our investigation to settings with $1$ (as used in Section~\ref{section-3.2}), $2$, and $4$ thinking nodes per trajectory. We control the number of thinking nodes by adjusting the thinking frequency. Specifically, we fix the maximum trajectory length to $25$ steps, and use a trigger interval of $9$ steps to obtain an average of $2$ thinking nodes per trajectory, and a trigger interval of $6$ steps to obtain an average of $4$ thinking nodes per trajectory.

\paragraph{Results.}
As shown in Table~\ref{fig: think-node-per-traj}, the single-thinking-node setting achieves the strongest overall performance. This result suggests that an excessive density of deep thinking within a single trajectory can destabilize policy optimization and reduce output reliability.

\begin{figure}[t]
\begin{center}
    \includegraphics[width=\linewidth]{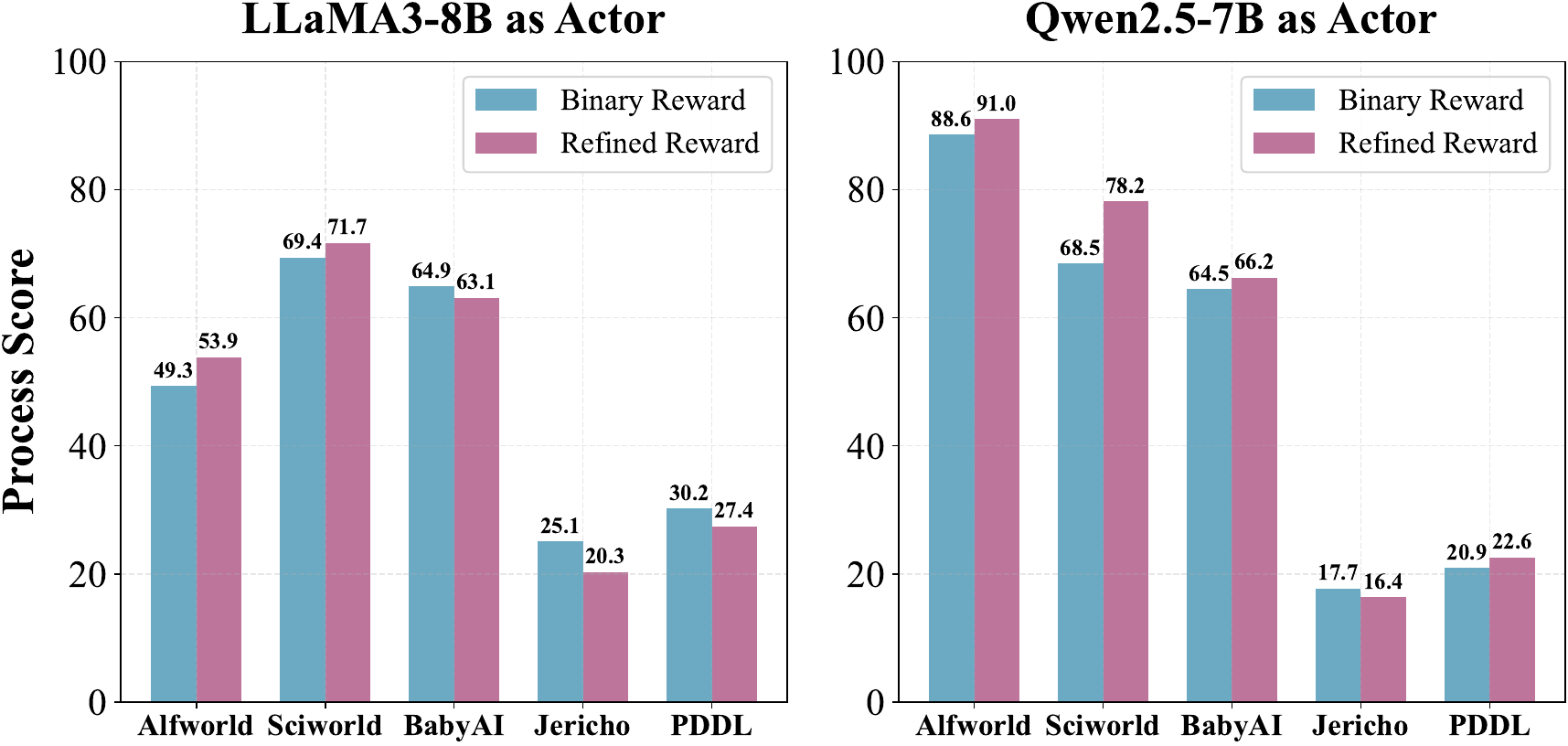}
\end{center}
\caption{Compare the usage of binary rewards and more refined rewards in the training of the thinker model.}
\label{fig: training-reward}
\end{figure}

% \subsection{Whether Using More Refined Rewards Will Be Better During Training?}
\subsection{Will More Refined Rewards Improve Training Performance?}
\label{section-4.8}

In this section, we discuss whether using more refined rewards to replace the binary rewards, which were used in our main experiment, will be better for the training of the thinker model. 
In this experiment, we introduced a step penalty of $- 0.05$ for each generation step taken before the first score increase. The longer it takes for the actor to achieve improvement, the more cumulative penalty is applied. Using this reward, we trained a new thinker. For a fair comparison, we also retrained our original thinker under the original binary reward, $0$ or $1$. Both models were trained on the same datasets, with identical hyperparameters, and for 400 steps, as the training curves show that both converge stably.
We then evaluated them using two base models as actors, \textit{Qwen2.5-7B-Instruct} and \textit{LLaMA3-8B-Instruct}, within our TTExplore framework. 

\paragraph{Results.} As shown in Figure~\ref{fig: training-reward}, the refined reward provides greater benefits on in-domain tasks, while its impact on out-of-domain tasks is limited. Consequently, we adopt the binary reward in our main experiments.

\begin{table}[t]
\centering
    \caption{The performance of different methods on Alfworld.}
    \setlength\tabcolsep{1mm}{
    \begin{tabular}{lcccc}
    \toprule
     Method & Know\% & Actor & Thinker & Alfworld \\
    \midrule
    Expel~\citep{zhao2024expel}     & 100\% & {\small \faSnowflake} prompt & -- & 41.04 \\
    KnowAgent~\citep{zhu2025knowagent} & 100\% & {\small \faFire} trained & -- & 75.37 \\
    WKM~\citep{qiao2024kwm}       & 100\% & {\small \faFire} trained & -- & 77.61 \\
    Knowself~\citep{qiao2025knowself}  & 15\% & {\small \faFire} trained & -- & \textbf{84.33} \\

    \midrule
    \multirow{3}{*}{TTExplore (LLaMA3)}
    & 0\% & {\small \faSnowflake} prompt & {\small \faSnowflake} prompt & 11.94 \\ 
    & 0\% & {\small \faSnowflake} prompt & {\small \faFire} trained & 32.08 \\ 
    & 0\% & {\small \faFire} trained & {\small \faFire} trained & 77.61 \\

    \midrule
    \multirow{3}{*}{TTExplore (Qwen2.5)}
    & 0\% & {\small \faSnowflake} prompt & {\small \faSnowflake} prompt & 57.46 \\
    & 0\% & {\small \faSnowflake} prompt & {\small \faFire} trained & 90.29 \\ 
    & 0\% & {\small \faFire} trained & {\small \faFire} trained & \textbf{97.76} \\
    \bottomrule
  \end{tabular}}
  \label{tab: more-baselines}
\end{table}

\subsection{Comparison of Offline-Guidance Methods}
\label{section-4.9}

In this section, we compare our test-time exploration approach with knowledge-augmented methods, including Expel~\citep{zhao2024expel}, KnowAgent~\citep{zhu2025knowagent}, KWM~\citep{qiao2024kwm}, and KnowSelf~\citep{qiao2025knowself}, which rely on offline-collected environment knowledge to enhance agent performance. 

\paragraph{Results.}
As shown in Table~\ref{tab: more-baselines}, the final column reports the success rate, and Know\% indicates the percentage of actions enhanced by external knowledge. By employing its \textit{Thinker} role to explore implicit environmental rules, TTExplore achieves performance comparable to that of offline knowledge‑guided methods. When paired with a strong backbone model (\textit{Qwen2.5‑7B‑Instruct}), our method achieves a success rate of $97.76\%$, setting a new SOTA result.

\section{Related Works}

Agents based on Large Language Models (LLMs) have shown strong performance across various scenarios~\citep{liu2024agentbench, chang2024agentboard}, including tool use, GUI navigation, and embodied tasks~\citep{cote2018textworld, wang2022scienceworld, shridhar2020alfworld}. Early approaches relied primarily on prompt-based methods~\citep{yao2023react, chen2023fireact} or framework design~\citep{lin2023swiftsage} to elicit effective behaviors. More recent studies focus on enhancing agent performance through two main directions: dedicated agent training and external knowledge injection.

Training-oriented approaches typically focus on collecting environments and tasks to strengthen agents’ planning and reasoning abilities. Representative works include AgentTuning~\citep{zeng2024agenttuning}, FireAct~\citep{chen2023fireact}, Agent-FLAN~\citep{chen2024agent-flan}, AgentOhana~\citep{zhang2024agentohana}, and AgentBANK~\citep{song2024agentbank}. However, such agents often struggle with out-of-domain generalization. To address this, some studies expand the diversity of training environments and tasks. For instance, AgentGen~\citep{hu2025agentgen} improves planning through synthesized tasks, while AgentRefine~\citep{fu2025agentrefine} enhances generalization by corrective data trajectories.

Alternatively, acquiring knowledge or experience about unknown environments has proven effective in boosting agent performance~\citep{xiao2023o3d, yang2024embodied, liu2025structured}. ExpeL~\citep{zhao2024expel} extracts rules from failed task trajectories to guide agents in new scenarios. Similarly, LWM~\citep{xiang2023lwm}, KnowAgent~\citep{zhu2025knowagent}, and KWM~\citep{qiao2024kwm} utilize pre-acquired knowledge bases to support better planning. KnowSelf~\citep{qiao2025knowself} advances this idea by enabling agents to actively seek environment-specific knowledge in critical steps. AgentRM~\citep{xia2025agentrm} uses an external reward model to guide the agent generation with beam search.

In summary, existing methods primarily rely on two strategies: \textbf{offline-memory}, where agents are exposed to more environments during training, and \textbf{offline-guidance}, which depends on extensive pre-collected knowledge. Both approaches, however, lack test-time exploration, which means acquiring and utilizing environmental knowledge dynamically during interaction. We argue that agents should instead emulate human-like behavior: engaging in test-time exploration, continuously interacting with their environment to gather information, distill experiences, infer underlying rules, and ultimately accomplish tasks autonomously.

\section{Conclusion}
We propose \textbf{TTExplore}, a framework that equips agents with periodic deep thinking to improve exploration and decision-making in unknown environments. By introducing a dedicated thinker trained via a stable pipeline, TTExplore enables agents to better infer implicit rules and plan effective actions. Experiments across five text-based embodied tasks demonstrate strong performance gains and cross-domain generalization, alleviating common issues such as repetitive behaviors and local exploration traps.

\begin{acks}
We sincerely thank all the anonymous reviewers
and (S)ACs for their constructive comments and
helpful suggestions. This work was supported by
The National Natural Science Foundation of China (No.\ 62376273 \& U2436209) and Beijing Natural Science Foundation (L253001).
\end{acks}

\bibliographystyle{ACM-Reference-Format}
\bibliography{sample_revise}

% \newpage

\appendix
\section{Method Details}
% 方法细节\prompt

\subsection{Prompts and Output Formats}
\label{append B1}

We show our prompts for the actor role and the thinker role in our framework, as shown in Figure~\ref{fig: actor-prompt} and Figure~\ref{fig: thinker-prompt}. The output format for the actor role is formula $\textless think\textgreater \dots \textless /think\textgreater \textless answer\textgreater \dots \textless /answer\textgreater $, which follows a ReAct-style~\citep{yao2023react}. The output format of the thinker role is formula $\textless deepthink\textgreater \dots \textless/deepthink\textgreater$. 

\begin{tcolorbox}[
    breakable, 
    title=Actor Prompt,
    colback=gray!5!white, 
    colframe=gray!75!black,
    fontupper=\normalsize,
    after skip=1.8mm
]
You are an Action Agent responsible for achieving a text-based task. 

Now you need to finish a text-based task in an environment with multi-turn interaction. 

Task Examples: \textcolor{blue}{[EXAMPLES]}

Task Actions: \textcolor{blue}{[ACTION SPACE]}

The Task: \textcolor{blue}{[TASK]}

Initial Observation: \textcolor{blue}{[INIT OBSERVATION]}

Attention:

1. You MUST provide your thought (one or two lines) before taking action.

2. You MUST issue only ONE action in each interaction stage.

Use the following format:
\begin{Verbatim}
<think> put your thought here </think>
<answer> put your action here </answer>
\end{Verbatim}
Please provide your response to the task following the format strictly.
\end{tcolorbox}
\captionof{figure}{The prompt for the actor role.}
\label{fig: actor-prompt}

\begin{tcolorbox}[
    breakable, 
    title=Thinker Prompt,
    colback=gray!5!white, 
    colframe=gray!75!black,
    fontupper=\normalsize,
    after skip=1.8mm % 直接减少框后间距
]
% \begin{Verbatim}[breaklines=true]
You are a Thinker Agent responsible for uncovering the implicit rules of the environment. You must analyze the history trajectory carefully and reason about any confusing feedback from the environment.

Here is the information about the task environment.

Task Actions: \textcolor{blue}{[ACTION SPACE]}

The Task: \textcolor{blue}{[TASK]}

Initial Observation: \textcolor{blue}{[INIT OBSERVATION]}

History Trajectory: \textcolor{blue}{[INTERACTION HISTORY]}

Attention:

1. If you think all the feedback in the history trajectory is reasonable, summarize the subgoals you have completed and provide your next plan.

2. If you find the environment's feedback in the latest steps confusing, think carefully about possible reasons. Do not assume the environment is erroneous; instead, consider what hidden rules could explain the observations.

3. For any uncertainties, try to formulate hypotheses and design plans to verify them.

Use the following format for your response:
\begin{Verbatim}
<deepthink> put your thought here </deepthink>
\end{Verbatim}
\end{tcolorbox}
\captionof{figure}{The prompt for the thinker role.}
\label{fig: thinker-prompt}

\subsection{The Training Pipeline for the Thinker Role}
\label{append B2}

In this section, we present our approach for training a professional thinker model based on a small backbone. We train our thinker model in two stages: supervised fine-tuning learning and reinforcement learning.

\paragraph{Supervised Fine-Tuning Stage}
% distillation
When analyzing environmental interaction trajectories, large models demonstrate stronger deep reasoning abilities than smaller ones. To bridge this gap, we apply supervised fine-tuning, enabling the small model to \textbf{quickly acquire the reasoning patterns of the larger model}. We assign a small model as the actor and a large model as the thinker, and use our TTExplore framework to generate task execution trajectories. Since the goal is to enhance the small model’s reasoning ability, we only extract the thinker’s outputs as training data. 

\paragraph{Reinforcement Learning Stage}
To further improve the thinker’s performance, we train it with \textbf{reinforcement learning} using an indirect reward. We first decompose complex tasks into sub-tasks and filter them by difficulty. In the sub-tasks division stage, we use a fine-tuned agent, Qwen2.5-Actor, which is based on Qwen2.5-7B-Instruct as the strong model. In the sub-tasks filtering stage, we have two thresholds, $x$ and $y$, which are mentioned in Section~\ref{section-3.2}. In our experiments, we set $x=5$ and $y=15$. We use a basic model, Qwen2.5-7B-Instruct, as the weak model. During the rollout stage, the trainable thinker generates multiple deep thoughts. In the reward computation stage, a fixed actor model executes subsequent trajectories based on the sampled thoughts, and the resulting task completion performance serves as the indirect reward signal. We update the thinker’s parameters using the GRPO algorithm~\citep{shao2024deepseekmath}.

\section{Experiment Details}
% 实验细节
\subsection{The Examples of the Deep Thinking Results from Different Models.}

In this section, we show some examples of the deep thinking results when using different models to play the role of thinker: our professional thinker model (Exp-Thinker), the Qwen2.5-72B-Instruct, and the Qwen3-8B (thinking mode).

\begin{tcolorbox}[
    breakable, 
    title=A Deep Thinking Result from Exp-Thinker,
    colback=gray!5!white, 
    colframe=gray!75!black,
    fontupper=\normalsize,
    after skip=1.8mm % 直接减少框后间距
]
\textbf{Deep Thinking Result:}

The repeated failure to execute the command "go to outside" suggests that the environment does not recognize this command or that the location "outside" is not directly accessible from the workshop. This could be due to several reasons:\newline
1. The command might need to be phrased differently.\newline
2. There might be a specific action required to move to the outside, such as opening a door or using a different command.\newline
3. The environment might require a more detailed command, such as specifying a door or a path to the outside.\newline
\newline
Given the initial observation, the only door mentioned is to the hallway, which is open. This might be a clue that the hallway is the next logical step to reach the outside. However, the task specifies that the animals are in the ``outside'' location, and it's possible that the hallway might lead to the outside.\newline
\newline
\textbf{Next Plan:}\newline
1. Go to the door to the hallway.\newline
2. Open the door if it is not already open.\newline
3. Go to the hallway.\newline
4. Look around to see if the outside is accessible from the hallway.\newline
5. If the outside is accessible, proceed to find the animals and focus on the one with the longest life span, then the one with the shortest life span.\newline
\newline
\textbf{Action to take:}\newline
- go to the door to the hallway\newline
- open the door to the hallway\newline
- go to the hallway\newline
- look around
\end{tcolorbox}
\captionof{figure}{An example of the deep thinking results produced by our professional thinker Exp-Thinker.}
\label{fig: exp-thinker}

\begin{tcolorbox}[
    breakable, 
    title=A Deep Thinking Result from Qwen2.5-72B-Instruct,
    colback=gray!5!white, 
    colframe=gray!75!black,
    fontupper=\normalsize,
    after skip=1.8mm % 直接减少框后间距
]
\textbf{Deep Thinking Result:}

The task is to put a clean plate in the countertop. The history trajectory shows that I have already taken a plate from cabinet 1 and placed it on countertop 1. However, the plate is not yet clean, which is a crucial part of the task. The sinkbasin 1 is likely the receptacle where I can clean the plate. 
\newline
\newline
\textbf{Subgoals completed:}\newline
1. Go to cabinet 1.\newline
2. Take plate 1 from cabinet 1.\newline
3. Go to countertop 1.\newline
4. Put plate 1 in/on countertop 1.\newline
\newline
\textbf{Next plan:}\newline
1. Go to sinkbasin 1.\newline
2. Clean plate 1 with sinkbasin 1.\newline
3. Go to countertop 1.\newline
4. Put the clean plate 1 in/on countertop 1.\newline
\newline
The feedback in the history trajectory is reasonable. The environment has provided clear and expected responses to the actions taken. There are no confusing elements in the feedback, and the actions align with the logical steps required to complete the task. The next step is to clean the plate, which I will do at the sinkbasin 1.
% \begin{Verbatim}</deepthink>\end{Verbatim}
\end{tcolorbox}
\captionof{figure}{An example of the deep thinking results produced by Qwen2.5-72B-Instruct.}
\label{fig: qwen2.5-72B-thinker}

\begin{tcolorbox}[
    breakable, 
    title=A Deep Thinking Result from Qwen3-8B,
    colback=gray!5!white, 
    colframe=gray!75!black,
    fontupper=\normalsize,
    after skip=1.8mm % 直接减少框后间距
]
% \begin{Verbatim}<think>\end{Verbatim}
\textcolor{gray}{Okay, let's see. The task is to put two soapbar in garbagecan. The initial observation shows various objects and receptacles. The history trajectory has the user opening cabinets 1, 2, and 3. Cabinet 1 has a cloth 1 and a soapbottle 1. Cabinet 2 has a candle 2 and toiletpaper 1. Cabinet 3 has a cloth 2.}

% \textcolor{gray}{Wait, the task is about soapbar, but in cabinet 1 there's a soapbottle. Maybe the soapbar is different from soapbottle? Or maybe the user needs to find soapbar somewhere else. The user hasn't found any soapbar yet. The cabinets 1-3 don't have soapbar. The other receptacles like countertop 1, sinkbasin 1 or 2, maybe those places could have soapbar? Or maybe the soapbar is in another cabinet, like cabinet 4? The user hasn't checked cabinet 4 yet.}

\textcolor{gray}{...(omit the thinking context for the page limitation)...}

\textbf{Deep Thinking Result:}
The task requires placing two soapbar objects into the garbagecan. The history trajectory shows the user has opened cabinets 1-3, which contain cloth, soapbottle, candle, and toiletpaper, but no soapbar. 
% The initial observation mentions a cabinet 4, which has not been explored yet. Since the task involves soapbar, and the only related object found is a soapbottle (not soapbar), the next logical step is to check cabinet 4. If soapbar is not there, other receptacles like countertop 1 or sinkbasins may need to be inspected. 
...(omit the context for the page limitation)...
\newline
The environment's feedback is consistent so far, so no confusion is detected. The subgoal of opening cabinets 1-3 is complete. The next plan is to go to cabinet 4 and open it to search for soapbar.
% \begin{Verbatim}</deepthink>\end{Verbatim}
\end{tcolorbox}
\captionof{figure}{An example of the deep thinking results produced by Qwen3-8B. We use its thinking mode, which means the model will give a long thinking in its default thinking mode first. After that, the model will give a deep thinking result depending on our requirements.}
\label{fig: qwen3-8B-thinker}

\subsection{Evaluate Our Framework On More Models}
We choose different models for the actor role and the thinker role. The results are shown in Table~\ref{tab: framework_results}, which demonstrate the effectiveness of our framework.
% The results in Table~\ref{tab: framework_results} demonstrate the effectiveness of our framework.

\begin{table}[tb]
  % \scriptsize
  % \small
  \centering
  \caption{The results of our framework on different models.}
  \setlength\tabcolsep{0.75mm}{
  \begin{tabular}{lcccccc}
    \toprule
    \textbf{Thinker} & Alfworld & Sciworld & BabyAI & Jericho & PDDL & \textbf{Mean}\\
    \midrule
    \multicolumn{7}{c}{\textit{GPT-4o-mini as Actor}} \\
    \midrule
    No & 47.63 & 86.23 & 67.23 & 20.72 & 44.65 & 53.29 \\
    GPT-4o-mini & 50.74 & 83.87 & 71.17 & 25.38 & 51.30 & 56.49 \\
    Qwen2.5-72B & 64.80 & 82.80 & 72.99 & 30.30 & 51.52 & \textbf{60.48} \\
    \midrule
    
    \multicolumn{7}{c}{\textit{LLaMA3-8B as Actor}} \\
    \midrule
    No & 32.15 & 20.31 & 42.44 & 14.23 & 29.94 & 27.81 \\
    LLaMA3-8B & 33.00 & 24.58 & 39.86 & 23.49 & 29.99 & 30.18 \\
    Qwen2.5-72B & 48.13 & 21.32 & 41.72 & 25.97 & 36.05 & \textbf{34.64} \\
    \midrule
    
    \multicolumn{7}{c}{\textit{Qwen2.5-7B as Actor}} \\
    \midrule
    No & 80.03 & 39.82 & 50.16 & 8.95 & 25.38 & 40.87 \\
    Qwen2.5-7B & 76.18 & 54.61 & 63.27 & 11.93 & 21.08 & 45.41 \\
    Qwen2.5-72B & 92.91 & 69.20 & 72.08 & 20.57 & 28.76 & \textbf{56.70} \\
    \midrule
    
    \multicolumn{7}{c}{\textit{Qwen2.5-72B as Actor}} \\ 
    \midrule
    No & 93.84 & 76.74 & 70.07 & 30.87 & 70.50 & 68.40 \\
    Qwen2.5-72B & 92.91 & 81.62 & 73.00 & 37.07 & 72.47 & \textbf{71.41} \\
    \bottomrule
  \end{tabular}}
  \label{tab: framework_results}
\end{table}

\section{The Usage of LLMs}
% 大模型使用声明
We use LLMs (e.g., ChatGPT and DeepSeek-V3.1) to help polish the paragraphs of our paper.

\end{document}